\documentclass[11pt,a4paper]{article}
\usepackage[dvipsnames]{xcolor}
\usepackage[hyperref]{emnlp-ijcnlp-2019}
\usepackage{times}
\usepackage{latexsym}
\usepackage{amssymb}
\usepackage{amsmath}
\usepackage{bm}
\usepackage{url}
\usepackage{graphicx}
\usepackage{multirow}
\usepackage{pifont}
\usepackage{soul}
\usepackage{xspace}
\usepackage{enumitem}

\aclfinalcopy 

\newcommand{\model}{KEAG\xspace}
\newcommand{\xmark}{\ding{55}}
\newcommand{\hlc}[2][cyan]{{%
    \colorlet{foo}{#1}%
    \sethlcolor{foo}\hl{#2}}%
}
\title{Incorporating External Knowledge into Machine Reading for Generative Question Answering}

\author{Bin Bi, Chen Wu, Ming Yan, Wei Wang, \\
  \textbf{Jiangnan Xia, Chenliang Li} \\
  Alibaba Group \\
  {\tt \{b.bi, wuchen.wc, ym119608, hebian.ww\}@alibaba-inc.com} \\
  {\tt \{jiangnan.xjn, lcl193798\}@alibaba-inc.com} \\}

\date{}

\begin{document}
\maketitle
\begin{abstract}
Commonsense and background knowledge is required for a QA model to answer many nontrivial questions. Different from existing work on knowledge-aware QA, we focus on a more challenging task of leveraging external knowledge to generate answers in natural language for a given question with context.

In this paper, we propose a new neural model, Knowledge-Enriched Answer Generator (\model), which is able to compose a natural answer by exploiting and aggregating evidence from all four information sources available: question, passage, vocabulary and knowledge. During the process of answer generation, \model adaptively determines when to utilize symbolic knowledge and which fact from the knowledge is useful. This allows the model to exploit external knowledge that is not explicitly stated in the given text, but that is relevant for generating an answer. The empirical study on public benchmark of answer generation demonstrates that \model improves answer quality over models without knowledge and existing knowledge-aware models, confirming its effectiveness in leveraging knowledge.
\end{abstract}

\section{Introduction}
Question Answering (QA) has come a long way from answer sentence selection, relational QA to machine reading comprehension. The next-generation QA systems can be envisioned as the ones which can read passages and write long and abstractive answers to questions. Different from extractive question answering, generative QA based on machine reading produces an answer in true natural language which does not have to be a sub-span in the given passage.

Most existing models, however, answer questions based on the content of given passages as the only information source. As a result, they may not be able to understand certain passages or to answer certain questions, due to the lack of commonsense and background knowledge, such as the knowledge about what concepts are expressed by the words being read (lexical knowledge), and what relations hold between these concepts (relational knowledge). As a simple illustration, given the passage:\\
\emph{State officials in Hawaii on Monday said they have once again checked and confirmed that President Barack Obama was born in Hawaii.}\\
to answer the question: \emph{Was Barack Obama born in the U.S.?}, one must know (among other things) that Hawaii is a state in the U.S., which is external knowledge not present in the text corpus.

Therefore, a QA model needs to be enriched with external knowledge properly to be able to answer many nontrivial questions. Such knowledge can be commonsense knowledge or factual background knowledge about entities and events that is not explicitly expressed but can be found in a knowledge base such as ConceptNet~\cite{Speer:2016}, Freebase~\cite{Tanon:2016} and domain-specific KBs collected by information extraction~\cite{Fader:2011,Mausam:2012}. Thus, we aim to design a neural model that encodes pre-selected knowledge relevant to given questions, and that learns to include the available knowledge as an enrichment to given textual information.

In this paper, we propose a new neural architecture, Knowledge-Enriched Answer Generator (\model), specifically designed
to generate natural answers with integration of external knowledge. \model is capable of leveraging symbolic knowledge from a knowledge base as it generates each word in an answer. In particular, we assume that each word is generated from one of the four information sources: 1. question, 2. passage, 3. vocabulary and 4. knowledge. Thus, we introduce the \emph{source selector}, a sentinel component in \model that allows flexibility in deciding which source to look to generate every answer word. This is crucial, since knowledge plays a role in certain parts of an answer, while in others text context should override the context-independent knowledge available in general KBs.

At each timestep, before generating an answer word, \model determines an information source. If the knowledge source is selected, the model extracts a set of facts that are potentially related to the given question and context. A stochastic fact selector with discrete latent variables then picks a fact based on its semantic relevance to the answer being generated. This enables \model to bring external knowledge into answer generation, and to generate words not present in the predefined vocabulary. By incorporating knowledge explicitly, \model can also provide evidence about the external knowledge used in the process of answer generation.

We introduce a new differentiable sampling-based method to learn the \model model in the presence of discrete latent variables. For empirical evaluation, we conduct experiments on the benchmark dataset of answer generation MARCO~\cite{Nguyen:2016}. The experimental results demonstrate that \model effectively leverages external knowledge from knowledge bases in generating natural answers. It achieves significant improvement over classic QA models that disregard knowledge, resulting in higher-quality answers.

\begin{figure*}[t]
    \centering
    \includegraphics[width=\linewidth]{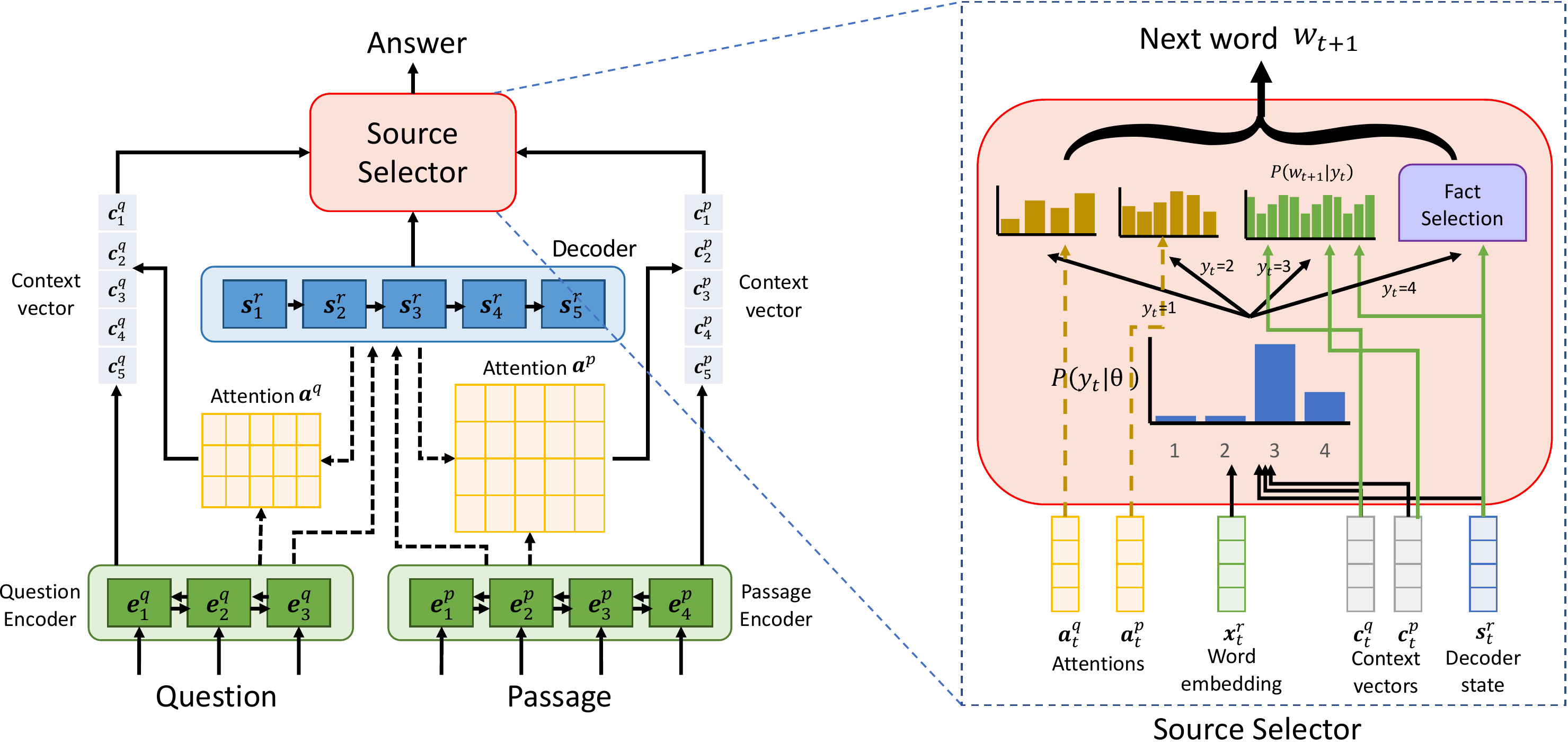}
    \caption{An overview of the architecture of \model (best viewed in color). A question and a passage both go through an extension of the sequence-to-sequence model. The outcomes are then fed into a source selector to generate a natural answer.}
    \label{fig:keag}
    \vspace{-10pt}
\end{figure*}

\section{Related Work}
\vspace{-5pt}
There have been several attempts at using machine reading to generate natural answers in the QA field. \citet{Tan:2018} took a generative approach where they added a decoder on top of their extractive model to leverage the extracted evidence for answer synthesis. However, this model still relies heavily on the extraction to perform the generation and thus needs to have start and end labels (a span) for every QA pair. \citet{Mitra:2017} proposed a seq2seq-based model that learns alignment between a question and passage words to produce rich question-aware passage representation by which it directly decodes an answer. \citet{Gao:2019} focused on product-aware answer generation based on large-scale unlabeled e-commerce reviews and product attributes. Furthermore, natural answer generation can be reformulated as query-focused summarization which is addressed by \citet{Nema:2017}.

The role of knowledge in certain types of QA tasks has been remarked on. \citet{Mihaylov:2018} showed improvements on a cloze-style task by incorporating commonsense knowledge via a context-to-commonsense attention. \citet{Zhong:2018} proposed commonsense-based pre-training to improve answer selection. \citet{Long:2017} made use of knowledge in the form of entity descriptions to predict missing entities in a given document. There have also been a few studies on incorporating knowledge into QA models without passage reading. GenQA~\cite{Yin:2016} combines knowledge retrieval and seq2seq learning to produce fluent answers, but it only deals with simple questions containing one single fact. COREQA~\cite{He:2017} extends it with a copy mechanism to learn to copy words from a given question. Moreover, \citet{Fu:2018} introduced a new attention mechanism that attends across the generated history and memory to explicitly avoid repetition, and incorporated knowledge to enrich generated answers.

Some work on knowledge-enhanced natural language (NLU) understanding can be adapted to the question answering task. CRWE~\cite{Weissenborn:2017} dynamically integrates background knowledge in a NLU model in the form of free-text statements, and yields refined word representations to a task-specific NLU architecture that reprocesses the task inputs with these representations. In contrast, KBLSTM~\cite{Yang:2017} leverages continuous representations of knowledge bases to enhance the learning of recurrent neural networks for machine reading. Furthermore, \citet{Bauer:2018} proposed MHPGM, a QA architecture that fills in the gaps of inference with commonsense knowledge. The model, however, does not allow an answer word to come directly from knowledge. We adapt these knowledge-enhanced NLU architectures to answer generation, as baselines for our experiments.

\section{Knowledge-aware Answer Generation}
\vspace{-5pt}
Knowledge-aware answer generation is a question answering paradigm, where a QA model is expected to generate an abstractive answer to a given question by leveraging both the contextual passage and external knowledge. More formally, given a knowledge base $\mathcal{K}$ and two sequences of input words: question $q=\{w_1^q, w_2^q, \dots, w_{N_q}^q\}$ and passage $p=\{w_1^p, w_2^p, \dots, w_{N_p}^p\}$, the answer generation model should produce a series of answer words $r=\{w_1^r, w_2^r, \dots, w_{N_r}^r\}$. The knowledge base $\mathcal{K}$ contains a set of facts, each of which is represented as a triple $f=(subject, relation, object)$ where $subject$ and $object$ can be multi-word expressions and $relation$ is a relation type, e.g., $(bridge, UsedFor, cross\ water)$.

\subsection{Knowledge-Enriched Answer Generator}
To address the answer generation problem, we propose a novel \model model which is able to compose a natural answer by recurrently selecting words at the decoding stage. Each of the words comes from one of the four sources: question $q$, passage $p$, global vocabulary $\mathcal{V}$, and knowledge $\mathcal{K}$. In particular, at every generation step, \model first determines which of the four sources to inspect based on the current state, and then generates a new word from the chosen source to make up a final answer.
An overview of the neural architecture of \model is depicted in Figure~\ref{fig:keag}.

\subsection{Sequence-to-sequence model}
\model is built upon an extension of the sequence-to-sequence attentional model~\cite{Bahdanau:2015,Nallapati:2016,See:2017}. The words of question $q$ and passage $p$ are fed one-by-one into two different encoders, respectively. Each of the two encoders, which are both bidirectional LSTMs, produces a sequence of encoder hidden states ($\mathbf{E}^q$ for question $q$, and $\mathbf{E}^p$ for passage $p$). In each timestep $t$, the decoder, which is a unidirectional LSTM, takes an answer word as input, and outputs a decoder hidden state $\mathbf{s}^r_t$.

We calculate attention distributions $\mathbf{a}^q_t$ and $\mathbf{a}^p_t$ on the question and the passage, respectively, as in~\cite{Bahdanau:2015}:
\begin{align}
\mathbf{a}^q_t=&\text{softmax}(\mathbf{g}^{q\intercal}\text{tanh}(\mathbf{W}^q\mathbf{E}^q+\mathbf{U}^q\mathbf{s}^r_t+\mathbf{b}^q)), \label{eqn:aq}\\ \nonumber
\mathbf{a}^p_t=&\text{softmax}(\\
&\mathbf{g}^{p\intercal}\text{tanh}(\mathbf{W}^p\mathbf{E}^p+\mathbf{U}^p\mathbf{s}^r_t+\mathbf{V}^p\mathbf{c}^q+\mathbf{b}^p)), \label{eqn:ap}
\end{align}
where $\mathbf{g}^q$, $\mathbf{W}^q$, $\mathbf{U}^q$, $\mathbf{b}^q$, $\mathbf{g}^p$, $\mathbf{W}^p$, $\mathbf{U}^p$ and $\mathbf{b}^p$ are learnable parameters. The attention distributions can be viewed as probability distributions over source words, which tells the decoder where to look to generate the next word. The coverage mechanism is added to the attentions to avoid generating repetitive text~\cite{See:2017}. In Equation~\ref{eqn:ap}, we introduce $\mathbf{c}^q$, a context vector for the question, to make the passage attention aware of the question context. $\mathbf{c}^q$ for the question and $\mathbf{c}^p$ for the passage are calculated as follows:
\begin{equation}
\mathbf{c}^q_t=\sum_i a^q_{ti}\cdot\mathbf{e}^q_i,\qquad \mathbf{c}^p_t=\sum_i a^p_{ti}\cdot\mathbf{e}^p_i,
\end{equation}
where $\mathbf{e}^q_i$ and $\mathbf{e}^p_i$ are an encoder hidden state for question $q$ and passage $p$, respectively. The context vectors ($\mathbf{c}^q_t$ and $\mathbf{c}^p_t$) together with the attention distributions ($\mathbf{a}^q_t$ and $\mathbf{a}^p_t$) and the decoder state ($\mathbf{s}^r_t$) will be used downstream to determine the next word in composing a final answer.

\section{Source Selector}
During the process of answer generation, in each timestep, \model starts with running a source selector to pick a word from one source of the question, the passage, the vocabulary and the knowledge. The right plate in Figure~\ref{fig:keag} illustrates how the source selector works in one timestep during decoding.

If the question source is selected in timestep $t$, \model picks a word according to the attention distribution $\mathbf{a}^q_t\in \mathbb{R}^{N_q}$ over question words (Equation~\ref{eqn:aq}), where $N_q$ denotes the number of distinct words in the question. Similarly, when the passage source is selected, the model picks a word from the attention distribution $\mathbf{a}^p_t\in \mathbb{R}^{N_p}$ over passage words (Equation~\ref{eqn:ap}), where $N_p$ denotes the number of distinct words in the passage. If the vocabulary is the source selected in timestep $t$, the new word comes from the conditional vocabulary distribution $P_v(w|\mathbf{c}^q_t,\mathbf{c}^p_t,\mathbf{s}^r_t)$ over all words in the vocabulary, which is obtained by:
\begin{equation}
P_v(w|\mathbf{c}^q_t,\mathbf{c}^p_t,\mathbf{s}^r_t)=\text{softmax}(\mathbf{W}^v\cdot[\mathbf{c}^q_t,\mathbf{c}^p_t,\mathbf{s}^r_t]+\mathbf{b}^v) \label{eqn:pv},
\end{equation}
where $\mathbf{c}^q_t$ and $\mathbf{c}^p_t$ are context vectors, and $\mathbf{s}^r_t$ is a decoder state. $\mathbf{W}^v$ and $\mathbf{b}^v$ are learnable parameters.

To determine which of the four sources a new word $w_{t+1}$ is selected from, we introduce a discrete latent variable $y_t\in \{1, 2, 3, 4\}$ as an indicator. When $y_t=1$ or $2$, the word $w_{t+1}$ is generated from the distribution $P(w_{t+1}|y_t)$ given by:
\begin{equation}
P(w_{t+1}|y_t)=
\begin{cases}
\sum_{i:w_i=w_{t+1}}a^q_{ti}& y_t=1\\
\sum_{i:w_i=w_{t+1}}a^p_{ti}& y_t=2.
\end{cases}
\end{equation}
If $y_t=3$, \model picks word $w_{t+1}$ according to the vocabulary distribution $P_v(w|\mathbf{c}^q_t,\mathbf{c}^p_t,\mathbf{s}^r_t)$ given in Equation~\ref{eqn:pv}. Otherwise, if $y_t=4$, the word $w_{t+1}$ comes from the fact selector, which will be described in the coming section.

\section{Knowledge Integration}
In order for \model to integrate external knowledge, we first extract related facts from the knowledge base in response to a given question, from which we then pick the most relevant fact that can be used for answer composition. In this section, we present the two modules for knowledge integration: \emph{related fact extraction} and \emph{fact selection}.

\subsection{Related Fact Extraction}
Due to the size of a knowledge base and the large amount of unnecessary information, we need an effective way of extracting a set of candidate facts which provide novel information while being related to a given question and passage.

For each instance $(q, p)$, we first extract facts with the subject or object that occurs in question $q$ or passage $p$. Scores are added to each extracted fact according to the following rules:
\begin{itemize}[noitemsep]
\item Score$+4$, if the subject occurs in $q$, \emph{and} the object occurs in $p$.
\item Score$+2$, if the subject \emph{and} the object both occur in $p$.
\item Score$+1$, if the subject occurs in $q$ \emph{or} $p$.
\end{itemize}
The scoring rules are set heuristically such that they model relative fact importance in different interactions. Next, we sort the fact triples in descending order of their scores, and take the top $N_f$ facts from the sorted list as the related facts for subsequent processing.

\subsection{Fact Selection}
\begin{figure}[t]
    \centering
    \includegraphics[width=0.6\linewidth]{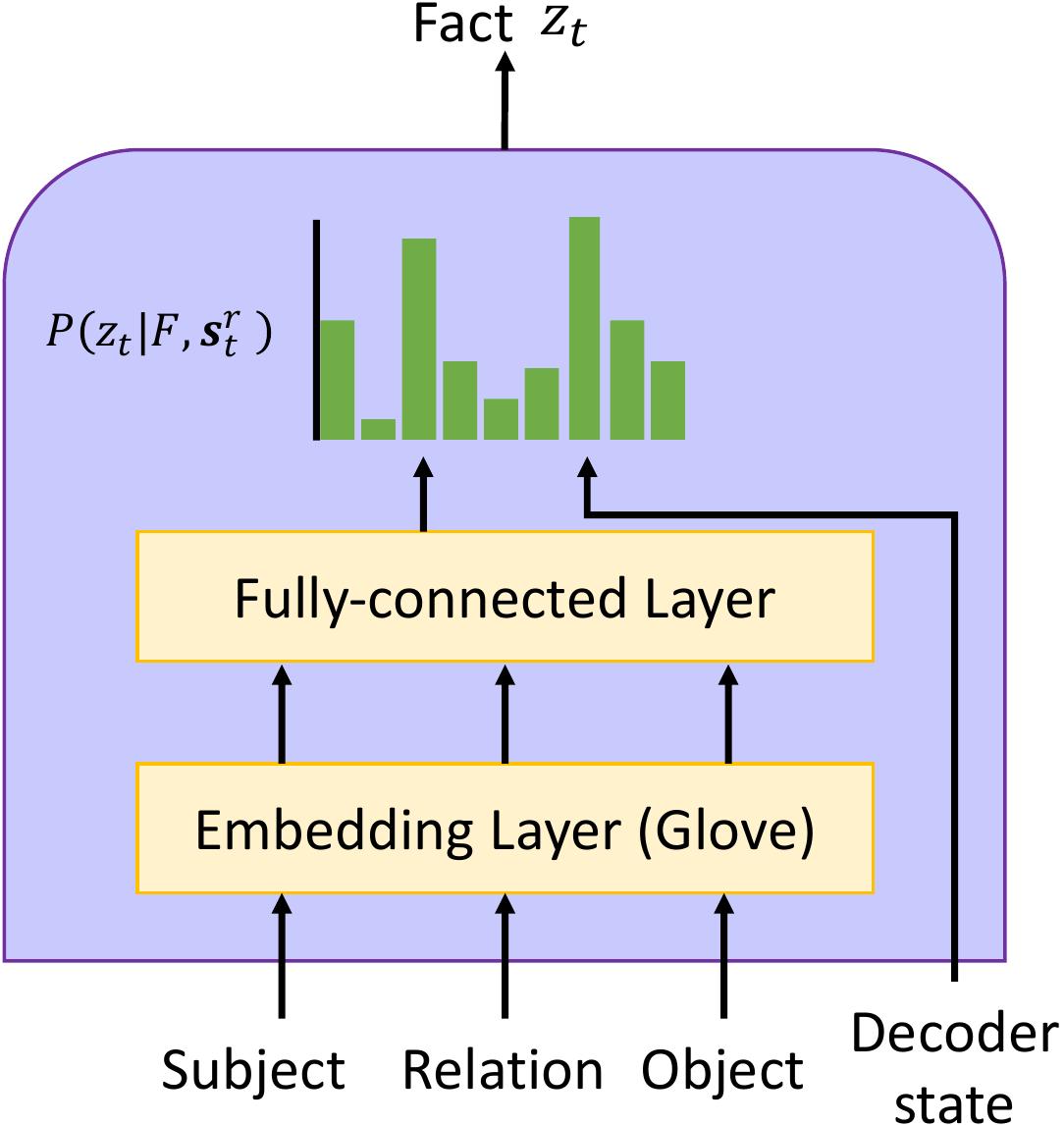}
    \caption{An overview of the fact selection module (best viewed in color)}
    \label{fig:factsel}
    \vspace{-10pt}
\end{figure}

Figure~\ref{fig:factsel} displays how a fact is selected from the set of related facts for answer completion. With the extracted knowledge, we first embed every related fact $f$ by concatenating the embeddings of the subject $\mathbf{e}^s$, the relation $\mathbf{e}^r$ and the object $\mathbf{e}^o$. The embeddings of subjects and objects are initialized with pre-trained GloVe vectors (and average pooling for multiple words), when the words are present in the vocabulary. The fact embedding is followed by a linear transformation to relate subject $\mathbf{e}^s$ to object $\mathbf{e}^o$ with relation $\mathbf{e}^r$:
\begin{equation}
\mathbf{f} = \mathbf{W}^e\cdot[\mathbf{e}^s,\mathbf{e}^r,\mathbf{e}^o]+\mathbf{b}^e.
\end{equation}
where $\mathbf{f}$ denotes fact representation, $[\cdot,\cdot]$ denotes vector concatenation, and $\mathbf{W}^e$ and $\mathbf{b}^e$ are learnable parameters. The set of all related fact representations $F=\{\mathbf{f}_1,\mathbf{f}_2,\dots,\mathbf{f}_{N_f}\}$ is considered to be a short-term memory of the knowledge base while answering questions on given passages.

To enrich \model with the facts collected from the knowledge base, we propose to complete an answer with the most relevant fact(s) whenever it is determined to resort to knowledge during the process of answer generation. The most relevant fact is selected from the related fact set $F$ based on the dynamic generation state. In this model, we introduce a discrete latent random variable $z_t\in [1, N_f]$ to explicitly indicate which fact is selected to be put into an answer in timestep $t$. The model selects a fact by sampling a $z_t$ from the discrete distribution $P(z_t | F, \mathbf{s}^r_t)$ given by:
\begin{equation}
P(z_t | \cdot)=\frac{1}{Z}\cdot\text{exp}(\mathbf{g}^{f\intercal}\text{tanh}(\mathbf{W}^f\mathbf{f}_{z_t}+\mathbf{U}^f\mathbf{s}^r_t+\mathbf{b}^f)),
\label{eqn:z_t}
\end{equation}
where $Z$ is the normalization term, $Z=\sum_{i=1}^{N_f} \text{exp}(\mathbf{g}^{f\intercal}\text{tanh}(\mathbf{W}^f\mathbf{f}_i+\mathbf{U}^f\mathbf{s}^r_t+\mathbf{b}^f))$, and $\mathbf{s}^r_t$ is the hidden state from the decoder in timestep $t$. $\mathbf{g}^f$, $\mathbf{W}^f$, $\mathbf{U}^f$ and $\mathbf{b}^f$ are learnable parameters.

The presence of discrete latent variables $\mathbf{z}$, however, presents a challenge to training the neural \model model, since the backpropagation algorithm, while enabling efficient computation of parameter gradients, does not apply to the non-differentiable layer introduced by the discrete variables. In particular, gradients cannot propagate through discrete samples from the categorical distribution $P(z_t | F, \mathbf{s}^r_t)$.

To address this problem, we create a differentiable estimator for discrete random variables with the Gumbel-Softmax trick~\cite{Jang:2017}. Specifically, we first compute the discrete distribution $P(z_t | F, \mathbf{s}^r_t)$ with class probabilities $\pi_1,\pi_2,\dots,\pi_{N_f}$ by Equation~\ref{eqn:z_t}. The Gumbel-Max trick~\cite{Gumbel:1954} allows to draw samples from the categorical distribution $P(z_t | F, \mathbf{s}^r_t)$ by calculating one\_hot$(\arg\max_i[g_i+\log\pi_i])$, where $g_1, g_2,\dots,g_{N_f}$ are i.i.d. samples drawn from the Gumbel$(0,1)$ distribution. For the inference of a discrete variable $z_t$, we approximate the Gumbel-Max trick by the continuous softmax function (in place of $\arg\max$) with temperature $\tau$ to generate a sample vector $\mathbf{\hat{z}}_t$:
\begin{equation}
\hat{z}_{ti}=\frac{\exp((\log(\pi_i)+g_i)/\tau)}{\sum_{j=1}^{N_f} \exp((\log(\pi_j)+g_j)/\tau)}.
\end{equation}
When $\tau$ approaches zero, the generated sample $\mathbf{\hat{z}}_t$ becomes a one-hot vector. $\tau$ is gradually annealed over the course of training.

This new differentiable estimator allows us to backpropagate through $z_t\sim P(z_t | F, \mathbf{s}^r_t)$ for gradient estimation of every single sample. The value of $z_t$ indicates a fact selected by the decoder in timestep $t$. When the next word is determined to come from knowledge, the model appends the object of the selected fact to the end of the answer being generated.

\section{Learning Model Parameters}
\vspace{-5pt}
To learn the parameters $\theta$ in \model with latent source indicators $\mathbf{y}$, we maximize the log-likelihood of words in all answers. For each answer, the log-likelihood of the words is given by:
\begin{align}
\vspace{-5pt}
&\log P(w_1^r, w_2^r, \dots, w_{N_r}^r|\theta)=\sum_{t=1}^{N_r} \log P(w_t^r|\theta) \nonumber \\
&=\sum_{t=1}^{N_r} \log \sum_{y_t=1}^4 P(w_{t+1}|y_t)P(y_t|\theta) \label{eqn:margin} \\
&\geq \sum_{t=1}^{N_r}\sum_{y_t=1}^4 P(y_t|\theta) \log P(w_{t+1}|y_t) \label{eqn:elbo} \\
&=\sum_{t=1}^{N_r} \mathbb{E}_{y_t|\theta}[\log P(w_{t+1}|y_t)], \label{eqn:expect}
\vspace{-5pt}
\end{align}
where the word likelihood at each timestep is obtained by marginalizing out the latent source variable $y_t$. Unfortunately, direct optimization of Equation~\ref{eqn:margin} is intractable, so we instead learn the objective function through optimizing its variational lower bound given in Equations~\ref{eqn:elbo} and \ref{eqn:expect}, obtained from Jensen's inequality.

To estimate the expectation in Equation~\ref{eqn:expect}, we use Monte Carlo sampling on the source selector variables $\mathbf{y}$ in the gradient computation. In particular, the Gumbel-Softmax trick is applied to generate discrete samples $\mathbf{\hat{y}}$ from the probability $P(y_t|\mathbf{c}^q_t, \mathbf{c}^p_t, \mathbf{s}^r_t, \mathbf{x}^r_t)$ given by:
\begin{equation}
P(y_t|\cdot)=\text{softmax}(\mathbf{W}^y\cdot[\mathbf{c}^q_t,\mathbf{c}^p_t,\mathbf{s}^r_t,\mathbf{x}^r_t]+\mathbf{b}^y),
\end{equation}
where $\mathbf{x}^r_t$ is the embedding of the answer word in timestep $t$, $\mathbf{W}^y$ and $\mathbf{b}^y$ are learnable parameters. The generated samples are fed to $\log P(w_{t+1}|y_t)$ to estimate the expectation.

\section{Experiments}
\vspace{-5pt}
We perform quantitative and qualitative analysis of \model through experiments. In our experiments, we also study the impact of the integrated knowledge and the ablations of the \model model. In addition, we illustrate how natural answers are generated by \model with the aid of external knowledge by analyzing a running example.

\subsection{Dataset and Evaluation Metrics}
Given our objective of generating natural answers by document reading, the MARCO dataset~\cite{Nguyen:2016} released by Microsoft is the best fit for benchmarking \model and other answer generation methods. We use the latest MARCO V2.1 dataset and focus on the ``\emph{Q\&A + Natural Language Generation}'' task in the evaluation, the goal of which is to provide the best answer available in natural language that could be used by a smart device / digital assistant.

In the MARCO dataset, the questions are user queries issued to the Bing search engine and the contextual passages are from real web documents. The data has been split into a training set (153,725 QA pairs), a dev set (12,467 QA pairs) and a test set (101,092 questions with unpublished answers). Since true answers are not available in the test set, we hold out the dev set for evaluation in our experiments, and test models for each question on its associated passages by concatenating them all together. We tune the hyper-parameters by cross-validation on the training set.

The answers are human-generated and not necessarily sub-spans of the passages, so the official evaluation tool of MARCO uses the metrics BLEU-1~\cite{Papineni:2002} and ROUGE-L~\cite{Lin:2004}. We use both metrics for our evaluation to measure the quality of generated answers against the ground truth.

For external knowledge, we use ConceptNet~\cite{Speer:2016}, one of the most widely used commonsense knowledge bases. Our \model is generic and thus can also be applied to other knowledge bases. ConceptNet is a semantic network representing words and phrases as well as the commonsense relationships between them. After filtering out non-English entities and relation types with few facts, we have 2,823,089 fact triples and 32 relation types for the model to consume.

\begin{table}[t]
\center
\begin{tabular}{ l | c | c}
\hline\hline
\textbf{Model} & \textbf{Rouge-L} & \textbf{Bleu-1}\\
\hline\hline
BiDAF & 19.42 & 13.03\\
\hline
BiDAF+Seq2Seq & 34.15 & 29.68\\
\hline
S-Net & 42.71 & 36.19\\
\hline
S-Net+Seq2Seq & 46.83 & 39.74\\
\hline
QFS & 40.58 & 39.96\\
\hline
VNET & 45.93 & 41.02\\
\hline
gQA & 45.75 & 41.10\\
\hline
\textbf{\model} & \textbf{51.68} & \textbf{45.97}\\
\hline\hline
\end{tabular}
\caption{Metrics of \model and QA models disregarding knowledge on the MARCO dataset.}
\label{table:dev}
\vspace{-10pt}
\end{table}

\subsection{Implementation Details}
In \model, we use 300-dimensional pre-trained \emph{Glove} word embeddings~\cite{Pennington:2014} for initialization with update during training. The dimension of hidden states is set to 256 for every LSTM. The fact representation $\mathbf{f}$ has $500$ dimensions. The maximum number of related facts $N_f$ is set to be 1000. We use a vocabulary of 50K words (filtered by frequency). Note that the source selector enables \model to handle out-of-vocabulary words by generating a word from given text or knowledge.

At both training and test stages, we truncate a passage to 800 words, and limit the length of an answer to 120 words. We train on a single Tesla M40 GPU with the batch size of 16. At test time, answers are generated using beam search with the beam size of 4.

\subsection{Model Comparisons}
Table~\ref{table:dev} compares \model with the following state-of-the-art extractive/generative QA models, which do not make use of external knowledge:
\begin{enumerate}
\itemsep0em
\item \textbf{BiDAF}~\cite{Seo:2017}: A multi-stage hierarchical process that represents the context at different levels of granularity, and using the bi-directional attention flow mechanism for answer extraction
\item \textbf{BiDAF+Seq2Seq}: A BiDAF model followed by an additional sequence-to-sequence model for answer generation
\item \textbf{S-Net}~\cite{Tan:2018}: An extraction-then-synthesis framework to synthesize answers from extracted evidences
\item \textbf{S-Net+Seq2Seq}: An S-Net model followed by an additional sequence-to-sequence model for answer generation
\item \textbf{QFS}~\cite{Nema:2017}: A model that adapts the query-focused summarization model to answer generation
\item \textbf{VNET}~\cite{Wang:2018}: An MRC model that enables answer candidates from different passages to verify each other based on their content representations
\item \textbf{gQA}~\cite{Mitra:2017}: A generative approach to question answering by incorporating the copying mechanism and the coverage vector
\end{enumerate}

Table~\ref{table:dev} shows the comparison of QA models in Rouge-L and Bleu-1. From the table we observe that abstractive QA models (e.g., \model) are consistently superior to extractive models (e.g., BiDAF) in answer quality. Therefore, abstractive QA models establish a strong base architecture to be enhanced with external knowledge, which motivates this work. Among the abstractive models, gQA can be viewed as a simplification of \model, which generates answer words from passages and the vocabulary without the use of knowledge. In addition, \model incorporates a stochastic source selector while gQA does not.  The result that \model significantly outperforms gQA demonstrates the effectiveness of \model's architecture and the benefit of knowledge integration.

\begin{table}[t]
\center
\begin{tabular}{ l | c | c}
\hline\hline
\textbf{Model} & \textbf{Rouge-L} & \textbf{Bleu-1}\\
\hline\hline
gQA w/ KBLSTM & 49.33 & 42.81\\
\hline
gQA w/ CRWE & 49.79 & 43.35\\
\hline
MHPGM & 50.51 & 44.73\\
\hline
\textbf{\model} & \textbf{51.68} & \textbf{45.97}\\
\hline\hline
\end{tabular}
\caption{Metrics of \model and knowledge-enriched QA models on the MARCO dataset.}
\label{table:knowledge}
\vspace{-5pt}
\end{table}

Table~\ref{table:knowledge} shows the metrics of \model in comparison to those of the following state-of-the-art QA models that are adapted to leveraging knowledge:
\begin{enumerate}
\itemsep0em
\item \textbf{gQA w/ KBLSTM}~\cite{Yang:2017}: KBLSTM is a neural model that leverages continuous representations of knowledge bases to enhance the learning of recurrent neural networks for machine reading. We plug it into gQA to make use of external knowledge for natural answer generation.
\item \textbf{gQA w/ CRWE}~\cite{Weissenborn:2017}: CRWE is a reading architecture with dynamic integration of background knowledge based on contextual refinement of word embeddings by leveraging supplementary knowledge. We extend gQA with the refined word embedding for this model.
\item \textbf{MHPGM}~\cite{Bauer:2018}: A multi-hop reasoning QA model which fills in the gaps of inference with commonsense knowledge.
\end{enumerate}

From Table~\ref{table:knowledge}, it can be clearly observed that \model performs best with the highest Rouge-L and Bleu-1 scores among the knowledge-enriched answer generation models. The major difference between \model and the other models is the way of incorporating external knowledge into a model. \emph{gQA w/ KBLSTM} and \emph{gQA w/ CRWE} extend gQA with the module that consumes knowledge, and MHPGM incorporates knowledge with selectively-gated attention while its decoder does not leverage words from knowledge in answer generation. Different from these models, \model utilizes two stochastic selectors to determine when to leverage knowledge and which fact to use. It brings additional gains in exploiting external knowledge to generate abstractive answers.

\begin{table}[t]
\center
\begin{tabular}{ l | c | c}
\hline\hline
\textbf{Model} & \textbf{Syntactic} & \textbf{Correct}\\
\hline\hline
gQA & 3.78 & 3.54\\
\hline
gQA w/ KBLSTM & 3.98 & 3.62\\
\hline
gQA w/ CRWE & 3.91 & 3.69\\
\hline
MHPGM & 4.10 & 3.81\\
\hline
\textbf{\model} & \textbf{4.18} & \textbf{4.03}\\
\hline\hline
\end{tabular}
\caption{Human evaluation of \model and state-of-the-art answer generation models. Scores range in $[1, 5]$.}
\label{table:man_dev}
\vspace{-7pt}
\end{table}

Since neither Rouge-L nor Bleu-1 can measure the quality of generated answers in terms of their correctness and accuracy, we also conduct human evaluation on Amazon Mechanical Turk. The evaluation assesses the answer quality on grammaticality and correctness. We randomly select 100 questions from the dev set, and ask turkers for ratings in a Likert scale ($\in [1, 5]$) on the generated answers.

Table~\ref{table:man_dev} reports the human evaluation scores of \model and state-of-the-art answer generation models. The \model model surpasses all the others in generating correct answers syntactically and substantively. In terms of syntactic correctness, \model and MHPGM both perform well thanks to their architectures of composing answer text and integrating knowledge. On the other hand, \model significantly outperforms all compared models in generating substantively correct answers, which demonstrates its power in exploiting external knowledge.

\subsection{Ablation Studies}
\vspace{-5pt}
\begin{table}[t]
\center
\begin{tabular}{l | c | c}
\hline\hline
\textbf{Ablation} & \textbf{Rouge-L} & \textbf{Bleu-1}\\
\hline\hline
\textbf{Full \model} & \textbf{51.68} & \textbf{45.97}\\
\hline
\xmark\ supplementary knowledge & 49.98 & 44.59\\
\hline
\xmark\ latent indicators $\mathbf{y}$ & 47.61 & 42.10\\
\hline
\xmark\ source selector & 38.33 & 36.75\\
\hline\hline
\end{tabular}
\caption{Ablation tests of \model.}
\label{table:ablation}
\vspace{-10pt}
\end{table}

We conduct ablation studies to assess the individual contribution of every component in \model. Table~\ref{table:ablation} reports the performance of the full \model model and its ablations.

We evaluate how much incorporating external knowledge as supplementary information contributes to natural answer generation by removing the supplementary knowledge and the corresponding fact selection module from \model's architecture. It can be seen that the knowledge component plays an important role in generating high-quality answers, with a drop to 49.98 on Rouge-L after the supplementary knowledge is removed.

To study the effect of our learning method, we further ablate the latent indicators $\mathbf{y}$, which leads to degradation to gQA except that the new model can select answer words from the question source while gQA cannot. Our learning method proves to be effective with a drop of about 5\% on Rouge-L and about 6\% on Bleu-1 after ablation.

Finally, for ablating the source selector, we have a new model that generates answer words from the vocabulary alone. It results in a significant drop to 38.33 on Rouge-L, confirming its effectiveness in generating natural answers.

\subsection{Visualization and Interpretation}
\vspace{-5pt}
\begin{table}[t]
\center
\begin{tabular}{ c | l }
\hline\hline
\multicolumn{2}{l}{\textbf{Question}}\\
\hline
\multicolumn{2}{l}{What's psychopathy?}\\
\hline\hline
\multicolumn{2}{l}{\textbf{Answer with source probabilities}}\\
\hline
\multirow{2}{*}{\textbf{\textcolor{blue}{Question src}}} & \hlc[cyan!61.68]{Psychopathy }\hlc[cyan!0.34]{is }\hlc[cyan!0.1]{a }\hlc[cyan!0.41]{personality }\\
& \hlc[cyan!0.12]{disorder}\hlc[cyan!0.07]{.}\\
\hline
\multirow{2}{*}{\textbf{\textcolor{Red}{Passage src}}} & \hlc[cyan!33.59]{Psychopathy }\hlc[cyan!1.21]{is }\hlc[cyan!3.51]{a }\hlc[cyan!11.49]{personality }\\
& \hlc[cyan!39.76]{disorder}\hlc[cyan!0.53]{.}\\
\hline
\multirow{2}{*}{\textbf{\textcolor{ForestGreen}{Vocabulary src}}} & \hlc[cyan!0.53]{Psychopathy }\hlc[cyan!98.44]{is }\hlc[cyan!96.34]{a }\hlc[cyan!4.33]{personality }\\
& \hlc[cyan!56.44]{disorder}\hlc[cyan!99.37]{.}\\
\hline
\multirow{2}{*}{\textbf{\textcolor{BurntOrange}{Knowledge src}}} & \hlc[cyan!4.2]{Psychopathy }\hlc[cyan!0.01]{is }\hlc[cyan!0.05]{a }\hlc[cyan!83.77]{personality }\\
& \hlc[cyan!3.67]{disorder}\hlc[cyan!0.03]{.}\\
\hline\hline
\multicolumn{2}{l}{\textbf{Answer colored by source}}\\
\hline
\multicolumn{2}{l}{\textcolor{blue}{Psychopathy} \textcolor{ForestGreen}{is a} \textcolor{BurntOrange}{personality} \textcolor{Red}{disorder}\textcolor{ForestGreen}{.}}\\
\hline\hline
\end{tabular}
\caption{Visualization of a sample QA pair and the source of individual words in the answer. The \textbf{Answer with source probabilities} section displays a heatmap on answer words selected from the question, passage, vocabulary and knowledge, respectively. A slot with a higher source probability is highlighted in darker cyan. The \textbf{Answer colored by source} section shows the answer in which every word is colored based on the source it was actually selected from. Words in \textcolor{blue}{blue} come from the question, \textcolor{Red}{red} from the passage, \textcolor{ForestGreen}{green} from the vocabulary, and \textcolor{BurntOrange}{orange} from the knowledge. The visualization is best viewed in color.}
\label{table:gen_ans}
\vspace{-1pt}
\end{table}

The source selector allows us to visualize how every word in an answer is generated from one of the sources of the question, passage, vocabulary and knowledge, which gives us insights about how \model works.

Table~\ref{table:gen_ans} visualizes a sample QA pair from \model and which source every word in the answer is selected from (indicated by the sample value of the source selector variable $y_t$). As exemplified in the table, the source distribution $P(y_t|\theta)$ varies over decoding timesteps. To answer the question, at each timestep, \model first selects a source based on the sample from  $P(y_t|\theta)$, followed by generating an answer word from the selected source. It is observed that the in the generated answer the keyword \emph{personality} comes from the knowledge source which relates psychopathy to personality. The answer word \emph{psychopathy} is selected from the question source, which leads to a well-formed answer with a complete sentence. Another keyword \emph{disorder}, on the other hand, comes from the passage source. This results from reading comprehension of the model on the passage. To generate a final answer in good form, \model picks the filler words \emph{is} and \emph{a} as well as the period ``\emph{.}'' from the vocabulary source. It makes the generated answer semantically correct and comprehensive.

\vspace{-5pt}
\section{Conclusion and Future Work}
\vspace{-5pt}
This paper presents a new neural model \model that is designed to bring symbolic knowledge from a knowledge base into abstractive answer generation. This architecture employs the source selector that allows for learning an appropriate tradeoff for blending external knowledge with information from textual context. The related fact extraction and stochastic fact selection modules are introduced to complete an answer with relevant facts.

This work opens up for deeper investigation of answer generation models in a targeted way, allowing us to investigate what knowledge sources are required for different domains. In future work, we will explore even tighter integration of symbolic knowledge and stronger reasoning methods.

\bibliography{emnlp-ijcnlp-2019}
\bibliographystyle{acl_natbib}

\end{document}